\documentclass{article}
\usepackage{ijcai24}

\usepackage{times}
\usepackage{soul}
\usepackage{url}
\usepackage[hidelinks]{hyperref}
\usepackage[utf8]{inputenc}
\usepackage[small]{caption}
\usepackage{graphicx}
\usepackage{amsmath}
\usepackage{amsthm}
\usepackage{booktabs}
\usepackage{algorithm}
\usepackage{algorithmic}
\usepackage[switch]{lineno}

\usepackage{multirow}
\usepackage{enumitem}

\newtheorem{definition}{Definition} 
\newtheorem{property}{Property}

\title{Review of Data-centric Time Series Analysis from Sample, Feature, and Period}

\author{
    Chenxi Sun$^{1,2,3}$
\and
Hongyan Li$^{1,2,3,4}$
\and
Yaliang Li$^5$
\and
Shenda Hong$^{6,7}$
\\
\affiliations
\large
$^1$National Key Laboratory of General Artificial Intelligence, Peking University \
$^2$Key Laboratory of Machine Perception (Ministry of Education), Peking University \
$^3$School of Intelligence Science and Technology, Peking University \
$^4$PKU-WUHAN Institute for Artificial Intelligence \
$^5$Alibaba Group \
$^6$National Institute of Health Data Science, Peking University \
$^7$Institute of Medical Technology, Health Science Center of Peking University
\\
\emails
\large
\{sun\_chenxi, leehy\}@pku.edu.cn,
yaliang.li@alibaba-inc.com, hongshenda@pku.edu.cn
}

\begin{document}

\maketitle

\begin{abstract}
Data is essential to performing time series analysis utilizing machine learning approaches, whether for classic models or today's large language models. A good time-series dataset is advantageous for the model's accuracy, robustness, and convergence, as well as task outcomes and costs. The emergence of data-centric AI represents a shift in the landscape from model refinement to prioritizing data quality. Even though time-series data processing methods frequently come up in a wide range of research fields, it hasn't been well investigated as a specific topic. To fill the gap, in this paper, we systematically review different data-centric methods in time series analysis, covering a wide range of research topics. Based on the time-series data characteristics at sample, feature, and period, we propose a taxonomy for the reviewed data selection methods. In addition to discussing and summarizing their characteristics, benefits, and drawbacks targeting time-series data, we also introduce the challenges and opportunities by proposing recommendations, open problems, and possible research topics.

\end{abstract}

\section{Introduction}

Time-Series (TS) data widely exists in real-world applications, such as industry, healthcare, finance, meteorology, etc. The growing interest in TS analysis, encompassing tasks like forecasting, classification, anomaly detection, and causality analysis, has garnered significant attention.

TS analysis has predominantly taken model-centric approaches, placing a primary emphasis on refining model designs to enhance performance using fixed datasets. The relevant models have evolved from traditional statistical approaches, including Auto-Regressive models (AR), Moving Average models (MA), and others, to sophisticated deep learning approaches like Recurrent Neural Networks (RNNs), Convolutional Neural Networks (CNNs), TS Transformers (TSTs), and their various adaptations. 

However, this model-centric approach often overlooks issues related to data quality, creating ambiguity regarding whether the enhancements in the model genuinely reflect its potential or if they are simply a result of overfitting to the dataset. Specifically, it is essential to note that when researching TS applications, tasks, or models, specific datasets must be provided, and the quality of the data directly influences the outcomes. The importance of data selection has persisted from classical models to contemporary Large Models (LM). It plays a crucial role in boosting the automation of classical models, contributing to the reduction of complexity and costs, and facilitating the adaptation of pre-trained Large Language Models (LLM) to TS data \cite{DBLP:journals/corr/abs-2308-08241}. 

Consequently, some emerging advocates have shifted their focus from model refinement to prioritizing data quality. But currently, the data-centric TS analysis has not been systematically investigated as a specialized field, even though it appears regularly across numerous study topics such as clustering, causality analysis \cite{DBLP:conf/ijcai/AssaadDG23}, few-shot learning \cite{DBLP:journals/corr/abs-2202-04769}, imbalanced classification \cite{DBLP:journals/ijon/ChouHZSSL20}, representation learning \cite{DBLP:journals/corr/abs-2308-01578}, curriculum learning \cite{sun2023curricular}, and Dimension reduction \cite{DBLP:journals/access/AshrafASCAIA23}, etc. 

\begin{figure}[!t]
\centering
\includegraphics[width=\linewidth]{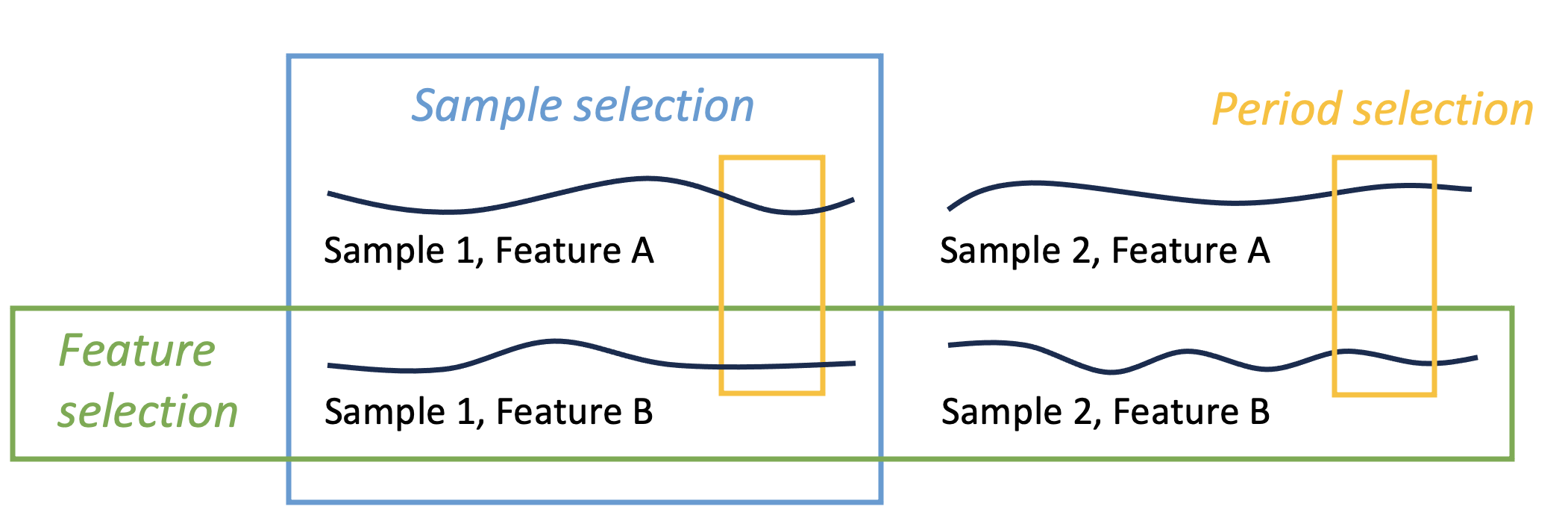}
\caption{Data-Centric Time Series Analysis: A case of time series selection for data quantity reduction from the perspectives of sample (blue), feature (green), and period (yellow).}
\label{fig:introduction} \vspace{-0.1cm}
\end{figure}

In this paper, to fill the aforementioned gaps, we review the utilization of data-related methods in TS analysis. We aim to analyze, summarize, and categorize various data selection methods based on the characteristics of TS data from three perspectives: sample (Section \ref{sec:sample}), feature (Section \ref{sec:feature}), and period (Section \ref{sec:length}). We discuss the benefits and drawbacks of related work (Table \ref{tab:summary}), provide conclusions and recommendations (Section \ref{sec:discussion}), and present the challenges and opportunities by proposing open problems and future work (Table \ref{tab:future}). 

We assert that data-centric approaches constitute a nascent frontier that complements existing efforts, and play an increasingly important role, not solely confined to TS analysis but also instrumental in diverse AI domains.

\section{Time-Series Data Characteristics} \label{sec:data}

\begin{definition}[Sample] \label{def:sample}
A TS sample is noted as $x=\{x^{d}_{i}\}_{i=1,d=1}^{T,D}$. It has $D$ variables (horizontal), $T$ time points (longitudinal) with value $x_{i}$ and possible timestamp record $t_{i}$, and the possible label information $c$. 
\end{definition}

\begin{definition}[Feature] \label{def:feature}
A multivariate TS $x=\{x^{d}_{i}\}_{i=1,d=1}^{T,D}$ has two types of features: 1) The intuitive features (original space) are reflected in its values in $D$ variables and its time information of records $t$. 2) The implied features (new space) are reflected in the relation between $D$ variables and new embeddings mapped in other spaces.
\end{definition}

\begin{definition}[Period] \label{def:length}
It represents a certain length on TS data. A full-length TS $x=\{x_{i}\}_{i=1}^{T}$ is represented as $T$ time steps or a time interval from $t_{1}$ to $t_{T}$. Its partial-length sample can be noted as some segmentations $x'=\{x_{j}\}_{j\geq 1}^{\leq T}$. 
\end{definition}

As shown in Figure \ref{fig:introduction}, in typical scenarios, data preprocessing is imperative after dataset collection, involving denoising and removing missing data. We can filter the data if the dataset is massive, and augment the data if it is too small. Then, when utilizing TS models, we can select appropriate features, windows, and learning orders to improve the model's convergence, accuracy, and robustness. 

Meanwhile, TS data in non-random systems often exhibits one or more of the following properties in two domains:

\begin{property}[Time domain properties] \label{prop:time}
1) Decomposability: $x=s+t+r$, $s$ is the seasonal term, $t$ is the trend term, and $r$ is the residual term; 2) Autocorrelation: $x_{i}=f(\{x_{j}\}_{j<i})$; 3) Variable correlation: $x^{d}=f(\{x^{j}\}_{j\neq d})$.
\end{property}

\begin{property}[Frequency domain properties]\label{prop:frequency}
1) Discrete Fourier Transform (DFT): The frequency spectrum is $F(\omega_{k})=\frac{1}{T}\sum_{i=1}^{T}x_{i}e^{-j\omega_{k}i}$. It can be decomposed into amplitude $A$ and phase spectra $\theta$ by $F(\omega_{k})=A(\omega_{k})\exp[j\theta(\omega_{k})]$ 2) Discrete Wavelet Transform (DWT): The approximation coefficients and detail coefficients of one-level DWT are $a_{n}=\sum_{k}h_{k}x_{2n-k}$ and $d_{n}=\sum_{k}g_{k}x_{2n-k}$. $h,g$ are the low-pass and high-pass filter coefficients.
\end{property}

\section{Sample Selection} \label{sec:sample}

This section covers the effects and requirements related to dataset size, quality, and learning order in various TS models, tasks, and scenarios. In summary, the primary research focus of the sample selection work outlined here is their concentration on existing models and their adaptation to the domain of TS analysis based on TS properties.

\begin{table}
    \centerline{
    \fontsize{7.5pt}{1}\selectfont
    \setlength{\tabcolsep}{0.5mm}{
    \begin{tabular}{lllll}
        \toprule
        Category &Topic &Task &Model &Dataset\footnotemark[1]\\
        \midrule
        \shortstack[l]{Sample\\\quad\\\quad\\\quad\\\quad\\\quad\\\quad\\\quad\\\quad\\\quad} &\shortstack[l]{Few-shot learning\\Imbalanced learning\\Transfer learning\\Meta-learning\\Federated learning\\Curriculum learning\\Continual learning} &\shortstack[l]{Clustering\\Generation\\Forecasting\\Classification\\One-class\\Multi-task\\\quad\\\quad} &\shortstack[l]{SVM\\CNN\\RNN\\TST\\SNN\\GAN\\DIFF\\\quad} &\shortstack[l]{Healthcare\\Industry\\Meteorology\\Finance\\Traffic\\Synthetic\\Archive} \\

        \midrule
        \shortstack[l]{Feature\\\quad\\\quad\\\quad\\\quad\\\quad\\\quad} &\shortstack[l]{Representation learning\\Contrastive learning\\Dimension reduction\\Causal learning\\Graph learning} &\shortstack[l]{Classification\\Forecasting\\Imputation\\Anomaly detection\\Causal inference} &\shortstack[l]{Tree\\GNN\\TCN\\ AE\\Attent.\\\quad} &\shortstack[l]{Healthcare\\Electricity\\Chaotic sys\\Archive\\\quad} \\

        \midrule
        \shortstack[l]{Period\\\quad\\\quad\\\quad\\\quad\\\quad\\\quad} &\shortstack[l]{Representation learning\\Dimension reduction\\Continual learning\\Online learning\\TST Design} &\shortstack[l]{Segmentation\\Staging, subtyping\\Anomaly detection\\Object localization\\Continuous classification} &\shortstack[l]{PCA\\DTW\\CNN\\TST\\\quad\\\quad} &\shortstack[l]{Activity\\Healthcare\\Archive\\\quad\\\quad\\\quad} \\
        \bottomrule
    \end{tabular}}}
    \caption{Research Topics, Models, and Datasets}
    \label{tab:reseach field}
\end{table}

\footnotetext[1]{\footnotesize{\url{https://github.com/SCXsunchenxi/ISMTS-Review}}}

\subsection{Data Filtering}

To ensure the quality of a large and noisy dataset, it becomes essential to extract high-quality data and perform data filtering. Most of the time, preprocessing steps, such as filtering out missing and noisy data, are employed after obtaining the TS dataset. While this method is straightforward, there is a risk of either insufficiently or excessively filtering the data.

Due to the varied evaluation criteria in different applications, the determination of what constitutes useful data varies. Consequently, most data filtering or sample reduction methods are tailored to specific tasks or fields. For instance, some methods focus on farthest boundary point estimation for support vector data description \cite{DBLP:journals/prl/AlamSANT20}, while others concentrate on common prototype generation for anomaly detection \cite{DBLP:conf/icml/LiCCWTZ23}. There are also methods designed for one-class classification such as distance-based representation \cite{DBLP:journals/pr/MauceriSM20} and local cluster balance \cite{DBLP:journals/eswa/HayashiCSFBCS24}.

\subsection{Data Augmentation}

The success of Deep Learning (DL) heavily relies on a large number of training data. This necessity becomes particularly crucial in the prevailing trend of LM. But in real-world applications, data scarcity is a common challenge, primarily manifesting in three situations: 1) The total samples are limited. That is a research topic in small sample learning, few-shot learning, zero-shot learning, meta-learning, etc. 2) The labeled samples is limited. That is a research topic in un-/self-supervised learning, positive unlabelled learning, active learning, etc. 3) The dataset is imbalanced. That is a research topic in imbalanced classification and regression, etc.

The problems in the above fields can be alleviated by increasing the number of total samples or samples in some specific classes. The main methods for accomplishing this involve data augmentation, generation, and synthesis. It is worth noting that their techniques are interconnected.


Unlike the static data, TS data possesses unique and intrinsic properties related to temporal dependency and time-frequency transformation. This complexity increases when modeling multivariate TS, where intricate dynamics among variables exist. Thus, the straightforward application of data augmentation methods from image and speech processing is not effective \cite{DBLP:conf/ijcai/Wen0YSGWX21}. Here, we summarize current data augmentation and expansion methods specifically for TS data into the following three categories:

\begin{itemize} [leftmargin=10pt,itemsep=0pt, topsep=1pt]
\item Basic methods in time domain and frequency domain.
\end{itemize}

Transformations in the time domain are relatively straightforward. Techniques involving noise injection directly manipulate the original TS data, including the addition of Gaussian noise, spikes, step-like, and slope-like trend patterns \cite{DBLP:conf/ijcai/wen19}. Time window-based methods, unique to TS, encompass window warping, flipping, cropping, slicing, jittering, perturbation, and their combinations.

Transformation in the frequency domain has inspired fewer augmentation methods. TS frequency can be decomposed into $A$ and $\theta$ as described in Property \ref{prop:frequency}. For randomly selected segments, $A$ and perturbations of $\theta$ are replaced with Gaussian noise with the original mean and variance, added by an extra zero-mean Gaussian noise, respectively \cite{DBLP:conf/MileTS/Gao20}. Additionally, the random phase shuffle can also be utilized after the amplitude-adjusted Fourier transform, which can approximately preserve the temporal correlation, amplitude distribution, and power spectra. Moreover, as TS has both time and frequency features, few methods use the time-frequency analysis \cite{DBLP:conf/interspeech/ParkCZCZCL19}.

\begin{itemize} [leftmargin=10pt,itemsep=0pt, topsep=1pt]
\item Generative models.
\end{itemize}

Statistical generative models intuitively describe TS dynamics, such as the mixture of Gaussian trees and the mixture autoregressive \cite{DBLP:journals/sadm/KangHL20}. These models rely on the autocorrelation in Property \ref{prop:time}, such conditional assumption will generate new sequences if initial values are perturbed.

Deep generative models, such as Generative Adversarial Networks (GANs) and Diffusion models (DIFF), have received significant attention in Computer Vision (CV). Their design needs to be adjusted for TS characteristics \cite{DBLP:journals/corr/abs-2305-00624}. For example, TimeGAN \cite{DBLP:conf/nips/YoonJS19} generates TS in various domains by a learned embedding space; TimeDiff \cite{DBLP:conf/icml/ShenK23} models TS by a diffusion model with future mixup and autoregressive initialization.

\begin{itemize} [leftmargin=10pt,itemsep=0pt, topsep=1pt]
\item Decomposition and synthesis.
\end{itemize}

Synthesis TS dataset can be obtained through some empirical formulas, such as Property \ref{prop:time} decomposability, Property \ref{prop:frequency} DFT, and time lag $x_{i}=\sum
_{j<i} \alpha_{j} x_{j}$. For example, Mackey-Glass system is $x_{i}=-bx_{i}+\frac{ax_{i-\tau}}{1+(x_{i-\tau})^{c}}$, where $\tau$ is the lag.

\subsection{Learning Order Arrangement}

Training models in a meaningful order can improve model performance over the standard approach based on random order. Curriculum Learning strategies (CL) train models from easier data to harder data, imitating the human curricula. We can summarize the easy-to-hard CL methods into predefined CL, where the difficulty measurer and the training scheduler are provided before training, and automatic CL, where the evaluation and scheduling dynamically change with training by self-paced learning, transfer teacher, or reinforcement learning strategies \cite{DBLP:conf/ijcai/PortelasCWHO20}. 

Although such an easy-to-use plug-in has demonstrated its power in improving the generalization capacity and convergence rate of various models in CV field \cite{DBLP:journals/pami/WangCZ22}, few of them are specifically designed for TS data. \cite{DBLP:journals/corr/abs-2311-13326} uses CL to smooth the noise during training for the control tasks over highly stochastic TS data. \cite{sun2023curricular} proposes a curricular and cyclical loss to train TS models based on the TS distribution changing dynamics.

\section{Feature Selection} \label{sec:feature}

TS data has many features, some of which are listed in Definition \ref{def:feature}. This section covers dimension reduction and feature augmentation methods. In summary, apart from conventional feature engineering techniques, most recent work has concentrated on deep features and representation learning.

\subsection{Feature Augmentation}\label{sec:feature augmentation}

Expanding features on existing ones can provide a comprehensive description. We summarize work into two categories:

\begin{itemize} [leftmargin=10pt,itemsep=0pt, topsep=1pt]
\item Add static features.
\end{itemize}

Including extra descriptions and statistical features in the input is a typical strategy for improving the performance of DL models. For instance, incorporating demographic information such as patient age and gender into medical TS data has been shown to improve diagnostic accuracy \cite{DBLP:journals/midm/SunDL21}. Descriptive statistics for TS encompass measures like mean, variance, skewness, and kurtosis, while trends involve moving averages, seasonality includes indices, and frequency involves histograms. Other characteristics, such as those related to stationarity testing, correlation, and autocorrelation analysis, can also be considered. These additional static features can be added to the input, concatenated to the hidden layer vectors, or integrated into the output \cite{DBLP:journals/nmi/ge20}.

\begin{itemize} [leftmargin=10pt,itemsep=0pt, topsep=1pt]
\item Add dynamic features.
\end{itemize}

TS has dynamics and dependencies over time development. The relations among variants may be changing. For this, graph-based methods have been utilized for TS analysis like GNN4TS \cite{DBLP:journals/corr/abs-2307-03759}. They can provide the interactive features \cite{DBLP:conf/dasfaa/DengCZZ21}, hypergraph-based local features \cite{DBLP:conf/ijcai/CaiSSZH022}, and spectral graph features \cite{DBLP:conf/ijcai/YuLYLHWL22}; The time information in TS data is likewise dynamic. For example, irregularly sampled TS has varying time intervals and lacks temporal alignment across dimensions. Such uneven time stamps can be encoded as the new dynamic features for DL models \cite{ijcai2021-414}. 

\begin{itemize} [leftmargin=10pt,itemsep=0pt, topsep=1pt]
\item Utilize representation learning.
\end{itemize}

The approach regarding the variant representation space of features can be applied in this context. Recently, universal representation learning for TS analysis has been popular. Most of them are unsupervised and learn discriminative feature representations from unlabeled data. This route also provides a foundation for implementing the general TS model. We summarize them into three categories:

Deep clustering methods allow the clustering-oriented iterative optimization, where clustering assignments can serve as self-supervised signals, thereby facilitating the self-learning process, such as CRLI \cite{DBLP:conf/aaai/MaCLC21} for incomplete TS, DTCR \cite{DBLP:conf/nips/MaZLC19} for temporal representations, IDFD \cite{DBLP:conf/iclr/TaoTN21} for spectral clustering, etc. They mainly focus on higher-level semantic information. 

Reconstruction methods minimize the difference between the reconstructed output and the raw input, ignoring insignificant features like noise. The related methods generally utilize the encoder-decoder architecture, such as RNN-based TimeNet and Transformer-based CRT \cite{DBLP:journals/corr/abs-2205-09928}.

Self-supervised learning methods design diverse pretext tasks that automatically generate useful supervised signals from the original data. \cite{DBLP:journals/corr/abs-2308-01578} classifies the TS self-supervised learning methods into three categories: adversarial methods, predictive methods, and contrastive methods. Among them, contrastive learning has been the subject of extensive research lately. They first generate positive and negative views of the anchor data and then employ the pretext task to bring similar pairs closer while pushing dissimilar pairs apart. Most related research is conducted in CV. For TS data, some efforts have been made to implement instance-level contrast \cite{DBLP:conf/iclr/WooLSKH22}, temporal-level contrast \cite{TLoss}, and clustering-level contrast. Meanwhile, the reconstruction methods introduced above can also be utilized as a pretext task for self-supervised learning to provide contextual information.

\subsection{Dimension Reduction} \label{sec:Reduce the Redundant Features}

Many real-world TS is characterized by high dimensionality. The dimensionality curse largely causes issues for learning approaches that can fail to capture the temporal dependencies. To address this, selecting a subset of features while preserving intrinsic properties becomes crucial. This not only reduces computational complexity but also prevents lower learning by eliminating irrelevant and redundant features. Here, we summarize the existing methods into two categories:

\begin{itemize} [leftmargin=10pt,itemsep=0pt, topsep=1pt]
\item Invariant representation space of features.
\end{itemize}

The approaches select $m$ from $n$ variables/dimensions ($m\leq n$) while maintaining the original feature distribution and numerical space. The straightforward method is the similarity measurement, which chooses one or more features from a set of similar features. The classic methods include Mahalanobis distance, DTW, and similarity-based clustering \cite{DBLP:conf/nips/MaZLC19}. But they may be sensitive to noise and have high complexity. Some recent methods consider the dynamic correlation among multiple variables, like according to the decomposition \cite{DBLP:journals/apin/ZhangZCHJXS23} and attention \cite{DBLP:journals/corr/abs-2311-16834}.

\begin{itemize} [leftmargin=10pt,itemsep=0pt, topsep=1pt]
\item Variant representation space of features.
\end{itemize}

Most algorithms change the original feature spaces, such as the well-known PCA, SVD, LE. For TS data, there are frequency PCA \cite{DBLP:journals/csda/Sundararajan21}, 2D SVD \cite{DBLP:journals/apin/GeWY23}, temporal extension LE  \cite{DBLP:journals/ijprai/HuangWY19}, etc. Additionally, the popular representation learning methods summarized in Section \ref{sec:feature augmentation} can also be used for dimension reduction purposes.

\begin{table*}
    \centering
    \resizebox{\textwidth}{!}{
    \begin{tabular}{llllllll}
        \toprule
        Category &Subcategory &Purpose &Method &Technology &Pro &Con &Paper\\
        \midrule
        \multirow{26}*{Sample} &\multirow{3}*{\shortstack[l]{Data\\filtering}} &\multirow{3}*{\shortstack[l]{Select high-\\quality data\\for training\\efficiency\\and model\\stability}}&\shortstack[l]{Filter missing\\and low-quality\\samples} &\shortstack[l]{Preprocessing:\\remove missing,\\noisy, etc} &\shortstack[l]{Necessary data\\processing steps,\\simple, intuitive} &\shortstack[l]{Need predefinition.\\Possible incomplete \\ or excessive filtering}  &\shortstack[l]{Filtering\\Sampling\\Denoising} \\
        
        \cline{4-8}
        & & &\shortstack[l]{Find \\important\\samples} &\shortstack[l]{Estimate farthest\\boundary points\\\quad} &\shortstack[l]{Enhance the\\classification\\boundary} &\shortstack[l]{Only targeting\\specific anomaly\\detection tasks} &\shortstack[l]{\cite{DBLP:journals/prl/AlamSANT20}\\ \cite{DBLP:journals/pr/MauceriSM20}\\\quad}\\ 
        
        \cline{4-8}
        & & &\shortstack[l]{Evaluate\\suitable\\samples} &\shortstack[l]{Clustering\\Prototyping\\\quad} &\shortstack[l]{Denoise and\\maintain data\\distribution} &\shortstack[l]{Make models less\\robust and prone\\to overfitting} &\shortstack[l]{
        \cite{DBLP:journals/eswa/HayashiCSFBCS24}\\
        \cite{DBLP:conf/icml/LiCCWTZ23}\\\quad}\\ 

            \cline{2-8}
           &\multirow{10}*{\shortstack[l]{Data\\augmenation}} &\multirow{10}*{\shortstack[l]{Increase the\\number of\\samples to\\reduce bias\\or distortion}} &\multirow{8}*{\shortstack[l]{Expand\\small\\datasets}} &\shortstack[l]{Time domain\\\quad\\\quad\\\quad\\\quad}  &\shortstack[l]{Simple, robust\\\quad\\\quad\\\quad\\\quad} &\shortstack[l]{Noises occur without\\prior knowledge\\\quad\\\quad} &\shortstack[l]{Window level\\ \cite{DBLP:journals/asc/GuoLQ23};\\Add noise\\ \cite{DBLP:conf/ijcai/wen19}} \\ 

            \cline{5-8}
            &&& &\shortstack[l]{Frequency domain\\\quad}  &\shortstack[l]{Simple, robust\\\quad} &\shortstack[l]{Noises occur without\\prior knowledge} &\shortstack[l]{Decompose\\\cite{DBLP:conf/MileTS/Gao20}}\\ 
            
            \cline{5-8}
            &&& &\shortstack[l]{Generate\\\quad\\\quad\\\quad\\\quad} &\shortstack[l]{Need less prior\\knowledge\\\quad\\\quad} &\shortstack[l]{Possible lack of\\temporal features\\\quad\\\quad} &\shortstack[l]{GAN-based\\\cite{DBLP:conf/nips/YoonJS19};\\Diffusion models\\\cite{DBLP:conf/icml/ShenK23}}\\ 

            \cline{5-8}
            &&& &\shortstack[l]{Synthetise\\\quad\\\quad} &\shortstack[l]{Controllable,\\flexible\\\quad} &\shortstack[l]{Need empirical\\formula\\\quad} &\shortstack[l]{Decompose expression\\\cite{DBLP:journals/corr/abs-2311-01933};\\Correlation expression}\\ 
            
            \cline{4-8}
            & & &\multirow{2}*{\shortstack[l]{Expand\\imbalanced\\-class datasets}}&\shortstack[l]{Oversampling} &\shortstack[l]{Simple, intuitive} &\shortstack[l]{Overfitting risk} &\shortstack[l]{\cite{DBLP:conf/ijcai/ZhaoLQWRL022}}\\ 

            \cline{5-8}
            & & & &\shortstack[l]{Loss\\\quad} &\shortstack[l]{Improved minority\\class prediction} &\shortstack[l]{Sensitivity to\\Hyperparameters} &\shortstack[l]{\cite{DBLP:journals/tkde/IrcioLMML23}\\\quad} \\ 
               
               \cline{2-8} 
               &\multirow{6}*{\shortstack[l]{Order\\arrangement}} &\multirow{3}*{\shortstack[l]{Make models \\learn data in\\ order to help\\convergence\\and generaliz-\\ation}} &\multirow{4}*{\shortstack[l]{Easy-to-hard\\curriculum}} &\shortstack[l]{Predefined\\\quad} &\shortstack[l]{No additional\\training complexity}  &\shortstack[l]{Sensitivity to manual\\design, potential bias} &\shortstack[l]{\cite{DBLP:journals/pami/WangCZ22}\\\quad}\\ 
                
                \cline{5-8}
                & & & &\shortstack[l]{Automatic\\\quad\\\quad\\\quad\\\quad\\\quad} &\shortstack[l]{Adaptive\\learning process\\\quad\\\quad\\\quad} &\shortstack[l]{Complexity overhead,\\training instability\\\quad\\\quad\\\quad} &\shortstack[l]{Self-paced learning;\\Transfer teacher;\\Reinforcement learning\\\cite{DBLP:conf/ijcai/PortelasCWHO20}}\\ 

               \cline{4-8}
                & & &\shortstack[l]{Other\\curriculum} &\shortstack[l]{Cyclical\\\quad} &\shortstack[l]{Targeting time-\\series data} &\shortstack[l]{Only for specific\\downstream tasks} &\shortstack[l]{\cite{sun2023curricular}\\\quad}\\
            
        \midrule
        \multirow{16}*{Feature} &\multirow{5}*{\shortstack[l]{Feature\\augmentation}}&\multirow{5}*{\shortstack[l]{Enrich the\\ information\\of original\\data}}&\shortstack[l]{Add static\\information} &\shortstack[l]{Add statistic,\\description} &\shortstack[l]{More\\comprehensive} &\shortstack[l]{Need prior\\knowledge} &\shortstack[l]{\cite{DBLP:journals/nmi/ge20}\\\cite{DBLP:conf/cinc/HongWZWSLX17}}\\ 

        \cline{4-8}
                & & &\multirow{2}*{\shortstack[l]{Add dynamic\\information}} &\shortstack[l]{Add time\\features} &\shortstack[l]{Model temporal\\irregular relations} &\shortstack[l]{Need additional\\timestamp records} &\shortstack[l]{\cite{ijcai2021-414}\\\cite{DBLP:journals/corr/abs-2010-12493}}\\ 

                \cline{5-8}
                & & & &\shortstack[l]{Add correlation\\features} &\shortstack[l]{Establish relations\\between variants} &\shortstack[l]{Possible redundancy\\and complexity} &\shortstack[l]{\cite{DBLP:conf/ijcai/YuLYLHWL22}\\
                \cite{DBLP:journals/corr/abs-2307-03759}}\\ 
                
        \cline{4-8}
                & &  &\shortstack[l]{Representation\\learning} &\shortstack[l]{ Encode to high-\\dimensional space}  &\shortstack[l]{ Deep feature\\discovery}  &\shortstack[l]{ Interpretability\\challenges} &\shortstack[l]{ \cite{DBLP:journals/corr/abs-2308-01578}\\\quad}\\
        
                \cline{2-8}
                &\multirow{8}*{\shortstack[l]{Dimension\\reduction}} &\multirow{8}*{\shortstack[l]{Reduce\\redundancy\\of high-\\dimensional\\data}} &\multirow{2}*{\shortstack[l]{Invariant\\feature domain}}&\shortstack[l]{Similarity-based\\\quad} &\shortstack[l]{Simple and\\explainable} &\shortstack[l]{Noise sensitivity,\\complexity} &\shortstack[l]{Clustering; Distance\\\cite{DBLP:journals/apin/ZhangZCHJXS23}} \\ 
                        
                \cline{5-8}
                & & & &\shortstack[l]{Attention-based\\\quad} &\shortstack[l]{Perceived sequent\\-ial features}   &\shortstack[l]{Still need to input\\all variables} &\shortstack[l]{End-to-end; Two-step\\\cite{DBLP:journals/corr/abs-2311-16834}}\\ 
 
                \cline{4-8}
                & & &\multirow{4}*{\shortstack[l]{Variant\\feature domain}} & \shortstack[l]{Time domain\\component analysis} &\shortstack[l]{multicollinearity\\mitigation} &\shortstack[l]{Outliers impact\\\quad} &\shortstack[l]{\cite{DBLP:journals/csda/Sundararajan21}\\\cite{DBLP:journals/apin/GeWY23}} \\ 
 
                \cline{5-8}
                & & & &\shortstack[l]{Frequency domain\\transformation} &\shortstack[l]{Global description\\\quad} &\shortstack[l]{Lost time\\information} &\shortstack[l]{\cite{DBLP:conf/icira/LiFHLZJZ22}\\\quad}\\

                \cline{5-8}
                & & & &\shortstack[l]{Shape domain\\extraction} &\shortstack[l]{Extract and show\\patterns} &\shortstack[l]{Complexity, hard for\\high-dimensionality} &\shortstack[l]{\cite{DBLP:conf/cikm/YamaguchiUK23}\\\quad}\\

        \midrule
        \multirow{14}*{Period} &\multirow{4}*{\shortstack[l]{Window\\size setting}} &\multirow{3}*{\shortstack[l]{Select the\\window\\size\\ of models}} &\shortstack[l]{Towards classi-\\cal models} &\shortstack[l]{Warping window\\\quad} &\shortstack[l]{More flexible\\and automated} &\shortstack[l]{Computational\\complexity} &\shortstack[l]{\cite{DBLP:conf/hais/Gomez-GonzalezC21}\\\quad}\\ 

        \cline{4-8}
                & & &\multirow{2}*{\shortstack[l]{Towards neural\\networks}} &\shortstack[l]{Input window\\size} &\shortstack[l]{Model-agnostic\\Task-agnostic} &\shortstack[l]{Need regressive\\hypothesis} &\shortstack[l]{\cite{DBLP:conf/kdd/Shima21}\\\quad}\\ 

                \cline{5-8}
                & & & &\shortstack[l]{Patching\\\quad} &\shortstack[l]{Adapt Transformers\\to time-series data} &\shortstack[l]{Need experience,\\not automatic} &\shortstack[l]{\cite{DBLP:conf/iclr/NieNSK23}\\\quad}\\ 
                
                \cline{2-8}
                &\multirow{3}*{\shortstack[l]{Data\\segmentation}} &\multirow{3}*{\shortstack[l]{Find\\meaningful\\segment\\sequences}} &\multirow{2}*{\shortstack[l]{Unsupervised}} &\shortstack[l]{Similarity-based\\\quad} &\shortstack[l]{Intuitive grouping,\\no prior assumption} &\shortstack[l]{Sensitive to distance\\metrics and outliers} &\shortstack[l]{\cite{DBLP:conf/icdm/BaiWLY020}\\\quad}\\ 

                \cline{5-8}
                & & & &\shortstack[l]{Model-based\\\quad} &\shortstack[l]{Model\\interpretability} &\shortstack[l]{Dependency on\\assumption} &\shortstack[l]{ARMA;\\GMM;HMM}\\ 
                
                \cline{4-8}
                & & &\shortstack[l]{Supervised\\\quad} &\shortstack[l]{Classification\\\quad} &\shortstack[l]{Handle complex\\patterns}&\shortstack[l]{Need task objectives\\or additional labels} &\shortstack[l]{\cite{DBLP:journals/datamine/ErmshausSL23}\\\quad}\\ 
                
                \cline{2-8}
                &\multirow{2}*{\shortstack[l]{Length\\reduction}} &\multirow{2}*{\shortstack[l]{Extract key\\sub-length\\sequences}} &\shortstack[l]{Longitudinal\\Dimension \\reduction} &\shortstack[l]{Classical dimension \\reduction methods\\\quad} &\shortstack[l]{ Can utilize\\different methods\\\quad} &\shortstack[l]{Only applicable\\to univariable data\\\quad} &\shortstack[l]{\cite{DBLP:journals/apin/SonbhadraAN23}\\\quad\\\quad}\\  

                \cline{4-8}
                & & &\shortstack[l]{Fragment\\extraction} &\shortstack[l]{Temporal pattern\\recognition} &\shortstack[l]{Interpretable, handle\\multivariable data} &\shortstack[l]{Need downstream\\task objective}
                &\shortstack[l]{\cite{DBLP:conf/aaltd/ErmshausSL22}\\\cite{10.3389/fphys.2021.811661}}
                \\
        \bottomrule
    \end{tabular}}
    \caption{Summary of Related Work on Data-centric Methods in Time Series Analysis}
    \label{tab:summary}
\end{table*}

\section{Period Selection} \label{sec:length}

TS data has the special length property as we described in Definition \ref{def:length}. TS often exhibit temporally repeated states. For example, human activity data is a series of repeated actions such as standing, walking, running, etc. Partitioning TS into a sequence of states can simplify and interpret the complex TS. The majority of TS period operations is dedicated to data segmentation: The unsupervised segmentation methods are usually distance-based or model-based \cite{DBLP:conf/icdm/BaiWLY020}; The supervised methods are supported by downstream objectives \cite{DBLP:journals/datamine/ErmshausSL23}. However, because segmentation is a specific task rather than the data processing and selection that we are interested in, we will not go into it in depth below.

\subsection{Window Size Setting}

Determining an appropriate window size for TS models can reduce model complexity and learn multi-span features. 

\begin{itemize} [leftmargin=10pt,itemsep=0pt, topsep=1pt]
\item Dynamic window size for classical models.
\end{itemize}

In DTW, obtaining the best performance requires setting its only parameter, the maximum amount of warping $w$; In model-based methods such as the Gaussian process, appropriate time-training intervals can extract unambiguous structural information from kernel hyperparameters. “The larger/smaller the better” needs to be permitted by the computational resources. Thus, in addition to reliable feature representation, many works focus on the complexity, such as utilizing semi-supervised dynamic optimization \cite{DBLP:conf/hais/Gomez-GonzalezC21}.

\begin{itemize} [leftmargin=10pt,itemsep=0pt, topsep=1pt]
\item Select the window size for neural networks.
\end{itemize}

Usually, except for some RNNs, when inputting TS data into a neural network, it is not a single point, but a subsequence, e.g. the sliding window for RNNs, the convolutional kernel for CNNs, and the patching for Transformers \cite{DBLP:conf/ijcai/WenZZCMY023}. These methods can be seen as the inherent structure of a model such as Temporal Convolutional Networks (TCNs), or as additional mechanisms before the model such as the external encoder \cite{DBLP:journals/midm/SunDL21}. 

In addition to the purpose of solving the online long term problem and end effect in online learning \cite{DBLP:journals/asc/GuoLQ23}, most studies start with a thorough modeling of the data from various levels, spans, and granularities, especially for RNN- or CNN-based models \cite{DBLP:conf/ijcai/ZhouHSWWLX19}. Currently, TS Transformers (TSTs) are the most popular model used for TS forecasting. PatchTST \cite{DBLP:conf/iclr/NieNSK23} emphasizes the importance of TS patching/tokenizing. Then self-attention focuses on the relations between local patterns. However, there is currently no agreement on the size of the patch. And the model-agnostic methods have not yet been implemented. 

Meanwhile, the window size setting may be related to specific datasets, a domain-agnostic method uses the moving average to set \cite{DBLP:conf/kdd/Shima21}, but it can only be considered semi-automatic as the requirements of a candidate set.

\begin{table*}[t]
    \centering
    \resizebox{\textwidth}{!}{
    \begin{tabular}{llll}
        \toprule
        Category &Key Technology and Domain &Scenario &Potential Topic\\
        \midrule
        \multirow{14}*{Sample} &\shortstack[l]{How to improve sample quality?\\Evaluation, data cleaning, large models} &\shortstack[l]{Large model and massive\\datasets, noisy dataset} &\shortstack[l]{$\bullet$ How to implement a data Juicer for TS-based\\\ \ \ large models?} \\[1ex]

        \cline{2-4}\\[-1.5ex]
       &\shortstack[l]{How to increase the number of samples?\\Few-shot learning, imbalanced\\classification, federated learning\\\quad\\\quad} &\shortstack[l]{Data collection difficulty,\\privacy. Small sample,\\skew dataset, cold start\\\quad\\\quad} &\shortstack[l]{$\bullet$ How to balance skewed datasets?\\ $\bullet$ How to create a larger synthetic dataset?\\ $\bullet$ How to generate finer TS data from different\\\ \ \ granularities? E.g. can we design Diffusion \\\ \ \ models combined with the progressive network?} \\[1ex]

        \cline{2-4}\\[-1.5ex]
       &\shortstack[l]{How to reduce the number of samples?\\Clustering, transfer learning\\\quad} &\shortstack[l]{ Noisy datasets, representative\\samples, explanations\\\quad} &\shortstack[l]{$\bullet$ How to find representative samples in one-class?\\$\bullet$ What is the law between data quantity and model\\\ \ \ quantity of TS-based large models?} \\[1ex]

        \cline{2-4}\\[-1.5ex]
       &\shortstack[l]{How to arrange the learning order?\\Curriculum learning, measurer, scheduler\\\quad\\\quad\\\quad} &\shortstack[l]{ High training costs, mode\\convergence difficulty,\\improve training benefits\\\quad} &\shortstack[l]{$\bullet$ How to achieve the efficient teacher-student for \\\ \ \ TS large models?\\ $\bullet$ How to solve the catastrophic forgetting problem\\\ \ \ in continuous learning by the learning order?} \\

       \midrule
       \multirow{5}*{Feature} &\shortstack[l]{How to select representative features?\\Correlation measurement, causal analysis} &\shortstack[l]{Explain, denoise, reduce\\complexity, improve robustness} &\shortstack[l]{$\bullet$ How to combine Granger causality analysis with\\\ \ \ dynamic hypergraphs?}\\[1ex]

       \cline{2-4}\\[-1.5ex]
       &\shortstack[l]{How to embed and represent features?\\Representation learning, embedding,\\unsupervised/self-supervised learning,\\contrastive learning} &\shortstack[l]{Unlabeled/partial-labeled data,\\the internal information is richer\\ than annotations.\\Subtyping, clustering} &\shortstack[l]{$\bullet$ How to achieve the TS universal representation\\\ \ \ as the foundation of the TS general model?\\ $\bullet$ How to achieve a flexible attention that spans\\\ \ \ horizontally and longitudinally?}\\

       \midrule
       \multirow{2}*{Period} &\shortstack[l]{How to choose the appropriate window\\size for TS model?\\\quad\\\quad\\\quad} &\shortstack[l]{Use models from other fields,\\ such as Transformers in NLP.\\ Segmentation, understanding\\\quad\\\quad} &\shortstack[l]{$\bullet$ How to design a flexible patching method for TS\\\ \ \ Transformers with dynamic window and spans?\\ $\bullet$ How to design a general window selection method\\\ \ \ for different TS models?}\\[1ex]

       \cline{2-4}\\[-1.5ex]
       &\shortstack[l]{How to extract key TS segments?\\Dimension reduction, pattern\\recognition, representation learning} &\shortstack[l]{Interpretation, semantic\\segmentation, turning point\\recognition, staging} &\shortstack[l]{$\bullet$ How to find key segments in long sequences\\ \ \ \ that can interpret the downstream labels (object\\\ \ \ detection on TS data)?}\\

        \bottomrule
    \end{tabular}}
    \caption{Potential Research Topics}
    \label{tab:future}
\end{table*}

\subsection{Subsequence Extraction}

The important insights from the TS data can be drawn by inspecting local substructures, and not the recordings as a whole \cite{DBLP:conf/aaltd/ErmshausSL22}. For example, ECG are medical TS that is repeated by normal or anomalous heartbeats within a longer sequence. As such, many SOTA methods characterize TS by inspecting local substructures \cite{10.3389/fphys.2021.811661}. And the method also needs to pay attention to the noise caused by non-stationary characteristics. Meanwhile, time pattern recognition is important for subtyping, staging, and interpretation. The time pattern is made up of key points from TS. For example, in medical mining tasks, it can discover disease stages, risk, and biomarkers \cite{SUN2024110075}.

TS length or span also can be treated as the dimension. The large sample length increases the computing complexities together with its innate problem of dimensionality curse, which can be solved through existing Dimension reduction methods introduced in Section \ref{sec:Reduce the Redundant Features}. But, towards the length reduction, to the best of our knowledge, only one publication has been reported \cite{DBLP:journals/apin/SonbhadraAN23}.

\section{Discussion}\label{sec:discussion}

We have summarized the data-related work across diverse fields (Table \ref{tab:reseach field}), We also distilled potential challenges and envisioned technological pathways (Table \ref{tab:future}), where we encapsulates the pivotal issues tailored to various scenarios, alongside the topics we put forward.

In this section, we first propose some open problems and share our reflections (Section \ref{sec:open problems}), then discuss some potential research topics requiring further exploration (Section \ref{sec:potential topics}). We hope that our newly proposed topics can give research inspiration to someone involved.

\subsection{Open Problems} \label{sec:open problems}

\begin{itemize} [leftmargin=10pt,itemsep=0pt, topsep=1pt]
\item More data or less data?
\end{itemize}

The exceptional performance of deep neural networks hinges significantly on the abundance of training data. However, TS data is more specialized, particularly in fields like medicine, where privacy and security are paramount concerns. The largest open dataset is typically less than 10GB, significantly smaller than those in Natural Language Processing (NLP) field \cite{zhou2023fits}. Hence, many effort is dedicated to enhancing the dataset.

However, in our engineering experience within healthcare \cite{HONG2024e4} and aerospace \cite{sun611}, what's surprising is that the challenge doesn't arise due to data scarcity; quite the opposite, data is often abundant. The challenge comes from a lack of well-managed methodologies for data accumulation, processing, and management, particularly when there's a shortage of individuals with a background in data science. Therefore, data organization, processing, and selection become increasingly crucial. Of course, in scenarios where data collection is notably challenging, such as with rare medical diseases, employing techniques like few-shot learning and imbalanced learning becomes justifiable.

On the other hand, TS is often derived from human-made processes which adhere to natural laws. E.g., the Earth's revolution and rotation create seasonal and diurnal patterns. It influences the rhythm and routines of human life and production, resulting in a consistent timing pattern observed across TS data in fields such as electricity, traffic, healthcare, and more, encompassing the timings of hours, days, weeks, months, and years. Because of this potential regularity or certain properties outlined in Section \ref{sec:data}, there might not be a necessity for an excessively large volume of data; it can be condensed. Especially in the current trend of LLM, once an adequate amount of data is acquired, researchers are contemplating methods to select high-quality data, optimize data utilization, and enhance overall efficiency \cite{chen2023datajuicer}.

Therefore, for TS data selection, we posit that a gradual process—initiating with expansion and subsequently transitioning to compaction—holds more validity. For data expansion, we can focus more on collection and management; For data compaction, we focus more on filtering and evaluation. 

\begin{itemize} [leftmargin=10pt,itemsep=0pt, topsep=1pt]
\item General large model or specialized model?
\end{itemize}

LMs stand as the current trend, and there have been initial endeavors in developing LM for TS data \cite{DBLP:journals/corr/abs-2310-10196}. Although they has demonstrated SOTA capabilities and, in some cases, exhibit few-shot or even zero-shot forecasting abilities, we can't confidently claim that they are genuinely general. Their models are typically trained on TS datasets specific to particular fields, where a forecasting dataset archive \cite{wu2023timesnet} is commonly used. We won't delve into the extent of this archive's representativeness at the moment, but it's worth noting that its validity can be easily questioned. In a general sense, the true general model would involve the capability to train on data from diverse fields concurrently and then deploy it across various domains. Retraining LLM-inspired TS models in a specific TS field can only produce a specific type of TST model, not a general one.

Meanwhile, from a data-centric perspective, unlike NLP's subjective text data that humans can perceive, TS is more objective, consolidating and uniformly describing them across various fields is challenging. These manifest in differences in sampling rates at sample level, the variability in variable number and correlations at feature level, and the potential for pattern alterations at period level. Although many TS data exhibit seasonality and trends, there's still a substantial amount of data from chaotic systems that are random or non-stationary. Moreover, most data selection methods are tailored for specific models and tasks, complicating efforts to standardize or unify them. For example, data augmentation methods applicable for classification may not be valid for anomaly detection \cite{DBLP:conf/ijcai/Wen0YSGWX21}. 

Hence, in the current technological landscape, it's more feasible to build distinct TS-LMs tailored to individual fields.

\subsection{Potential Topics} \label{sec:potential topics}

We provide many potential research topics as listed in the last column of Table \ref{tab:future}. Here we discuss some of them in detail:

\begin{itemize} [leftmargin=10pt,itemsep=0pt, topsep=1pt]
\item Skewed, imbalanced, or long-tailed distributed data issues.
\end{itemize}

In most practical applications where data collection is feasible, skewed data issues tend to be more severe than small samples. In electronic medical records, typical cases make up a significant portion, whereas rare or orphan diseases, despite receiving heightened attention, constitute a very small fraction. Atrial fibrillation records account for less than 0.5\% in ECG \cite{DBLP:journals/ijon/ChouHZSSL20}. In specific tasks, beyond the prevalent challenges of imbalanced classification problems, it's also crucial to note the significance of imbalanced or long-tailed regression \cite{DBLP:conf/aistats/StocksiekerPC23}.

\begin{itemize} [leftmargin=10pt,itemsep=0pt, topsep=1pt]
\item The scaling law between the number of TS data and that of TS model parameters.
\end{itemize}

To attain improved accuracy or and utilize continuously collected data, it might be necessary to expand both the training dataset and the models. In NLP, studies have investigated the scaling relationship between the amount of text data and the parameters within LLMs, drawing conclusions that align with power law \cite{DBLP:journals/corr/abs-2001-08361}. But this conclusion might not be entirely applicable to the TS data analysis. Hence, investigating the TS scaling law and crafting dedicated data selection methods like data-juicer \cite{chen2023datajuicer} is worthwhile for future research.

\begin{itemize} [leftmargin=10pt,itemsep=0pt, topsep=1pt]
\item Address multiple problems uniformly through the sample-level data-augmented generation.
\end{itemize}

Viewing forecasting, classification, imputation, and similar tasks as challenges associated with generation can offer unified strategies. Considering the entire sample can simplify the need to account for sequential dependencies, feature correlations, irregular time, and other intricate details. Diffusion models gradually climbed the ranks among SOTA methods for TS data \cite{DBLP:journals/corr/abs-2209-00796}. Despite all this, there's still a scarcity of well-established research on frequency domain diffusion and progressive models within TS analysis.

\begin{itemize} [leftmargin=10pt,itemsep=0pt, topsep=1pt]
\item 
Approach data processing and selection by considering relationships dynamically.
\end{itemize}

TS data evolves dynamically, with key features and segments undergoing continual changes at various intervals. For instance, the concept of dynamic Granger causality highlights this evolving nature \cite{sun611}. Thus, akin to online learning and continual learning paradigms \cite{sunpatterns}, data selection at both the feature and length levels could be more dynamic, immediate, and real-time-oriented. This suggests that a unified structure could be established by integrating an encoder as a selector positioned ahead of the TS model, e.g. it can complete dynamic patching for TST. 

\begin{itemize} [leftmargin=10pt,itemsep=0pt, topsep=1pt]
\item Utilize other data structures.
\end{itemize}

TS data can be roughly conceptualized with three sampling rate categories: low, medium, and high. For low-frequency data, like diagnosis events, they can be modeled in a tabular data format; For medium-frequency data, like vital signs, they can be structured using graphs and hypergraphs that emphasize feature correlation; For high-frequency data, like ECG physiological signals, they can be converted into images. This approach broadens the scope of available data selection methods, allowing for the utilization of techniques from diverse fields, not confined solely to temporal structures.

\section{Conclusion}

This paper is the first systematic review to investigate the data-centric methods for TS analysis as a specialized topic. We underscore the significance of data management and organization within the context of current technological trends. We acknowledge the significance of unified general models for TS data but emphasize that currently, the domain-specific general models and corresponding data selection methods are more practical and effective. We recommend to select TS data from all perspectives of sample, feature, and period. In the future, it is also necessary to study TS data-centric methods for data labeling, model interpretability and fairness, prompt engineering, and proposing new datasets or benchmarks.

\section*{Acknowledgments}
This work is supported by National Natural Science Foundation of China (No.62172018, No.62102008) and Wuhan East Lake High-Tech Development Zone National Comprehensive Experimental Base for Governance of Intelligent Society.

\bibliographystyle{named}
\bibliography{references}

\end{document}